\documentclass{article} 
\usepackage{iclr2023_conference,times}

\usepackage[utf8]{inputenc}
\usepackage{graphicx}
\usepackage{tcolorbox}
\usepackage[caption=false]{subfig}
\usepackage{amsmath}
\usepackage{amssymb,amsfonts}
\usepackage{comment}
\usepackage{color}
\usepackage{soulutf8}
\usepackage{url}
\usepackage{multirow}
\usepackage{siunitx}
\usepackage{booktabs}
\usepackage{multicol}
\usepackage{mathtools}
\usepackage{algorithm,algorithmic}
\usepackage{bbm}
\usepackage{cleveref}
\usepackage{cite}
\usepackage{wrapfig} 

\usepackage{caption}

\newcommand{\cfd}[1]{{\color{orange} #1}} 

\DeclareMathOperator*{\argmax}{argmax} 

\title{Towards Explainable Land Cover Mapping: a Counterfactual-based Strategy}

\author{Cassio F.~Dantas${}^{\dagger\ast}$, Diego Marcos${}^{\ddag\ast}$, Dino Ienco${}^{\dagger\ast}$ \\
        ${}^\dagger$UMR TETIS, INRAE, ${}^\ddag$Inria,
        ${}^\ast$University of Montpellier, France \\
        \texttt{\{cassio.fraga-dantas,dino.ienco\}@inrae.fr}, 
        \texttt{diego.marcos@inria.fr}
}

\iclrfinalcopy 
\begin{document}

\maketitle

\begin{abstract}
    Counterfactual explanations are an emerging tool to enhance interpretability of deep learning models. Given a sample, these methods seek to find and display to the user similar samples across the decision boundary.
    In this paper, we propose a generative adversarial counterfactual approach for satellite image time series in a multi-class setting for the land cover classification task. One of the distinctive features of the proposed approach is the lack of prior assumption on the targeted class for a given counterfactual explanation. This inherent flexibility allows for the discovery of interesting information on the relationship between land cover classes. The other feature consists of encouraging the counterfactual to differ from the original sample only in a small and compact temporal segment. These time-contiguous perturbations allow for a much sparser and, thus, interpretable solution. Furthermore, plausibility/realism of the generated  counterfactual explanations is enforced via the proposed adversarial learning strategy.
\end{abstract}

\section{Introduction}
Deep learning techniques have gained widespread popularity in the remote sensing field due to impressive results on a variety of tasks such as image super-resolution, image restoration, biophysical variables estimation and land cover classification from satellite image time series (SITS) data~\citep{YUAN2020111716}.
Of particular importance, this last task provides useful knowledge to support many downstream geospatial analyses~\citep{IngladaVATMR17}. Despite the high performances achieved by recent deep learning frameworks on this task, they remain black-box models with limited understanding on their internal behavior.
Due to this limitation, there is a growing need for improving the interpretability of deep learning models in remote sensing with the objective to raise up their acceptability and usefulness, as their decision-making processes are often not transparent~\citep{Adadi2018,Guidotti2019,Arrieta2020}. Counterfactual explanation methods have recently received increasing attention as a means to provide some level of interpretability~\citep{Wachter2017,verma2020counterfactual,Guidotti2022} to these black-box models. Counterfactual explanations aim to describe the behaviour of a model by providing minimal changes to the input data that would result in realistic samples that result in the model predicting a different class.

For these perturbations to be more easily interpretable it is desirable that they are sparse and that they can be identified with some semantic element of the input data. In the case of time series, this would require to perturb a short and contiguous section of the timeline~\citep{Delaney2021}.
Most papers on counterfactual explanations focus on image data, while much fewer concentrate on time series \citep{Delaney2021, Li2022, Ates2021, Guidotti2020, Lang2022, VanLooveren2021, FilaliBoubrahimi2022}.
To the best of our knowledge, this is the first paper focusing more specifically on counterfactuals for remote sensing time series data.
The proposed approach generates counterfactual explanations that are plausible (i.e. belong as much as possible to the data distribution) and close to the original data (modifying only a limited and contiguous set of time entries by a small amount). 
Finally, it is not necessary to pre-determine a target class for the generated counterfactual.

\paragraph*{Paper outline} In \Cref{sec:data} we describe the considered study case with the associated remote sensing data. After detailing the proposed method in \Cref{sec:method}, we present the experimental results in \Cref{sec:results}. Concluding remarks and future works are outlined in \Cref{sec:conclusion}.

\section{Study Area} \label{sec:data}
\begin{figure}[t]
\begin{minipage}{\textwidth}
  \begin{minipage}[b]{0.66\textwidth}
    \vspace{0cm}
    \centering
    \includegraphics[width=.99\columnwidth]{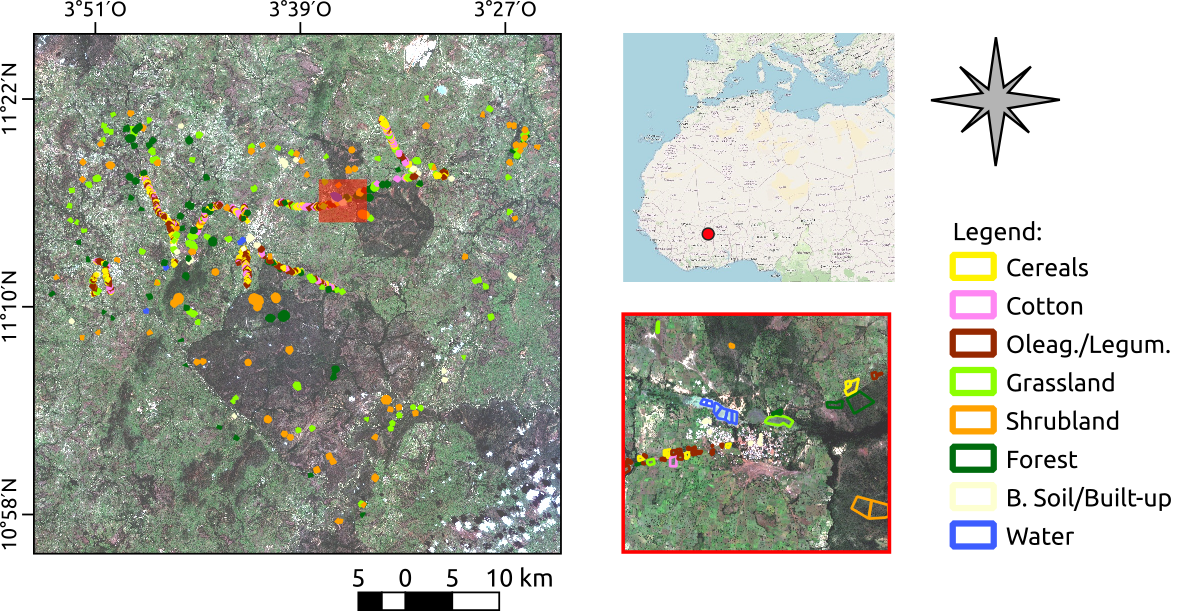}
    \captionof{figure}{Study site location and corresponding ground truth.\label{fig:loc-koumbia}}
  \end{minipage}%
  \hfill
  \begin{minipage}[b]{0.33\textwidth}
    \vspace{0cm}
    \begin{tabular}{ccc}
      \toprule
      Class &  Label & Pixels \\
      \cmidrule(lr){1-1} \cmidrule(lr){2-2} \cmidrule(lr){3-3}
      1 &  \emph{Cereals} & 9\,731 \\
      2 & \emph{Cotton} & 6\,971 \\
      3 & \emph{Oleaginous} & 7\,950 \\
      4	& \emph{Grassland} & 12\,998 \\
      5	& \emph{Shrubland} & 22\,546 \\
      6	& \emph{Forest} & 17\,435 \\
      7	& \emph{Bare soil} & 1\,125 \\
      8	& \emph{Water} & 1\,205 \\
      \cmidrule(lr){1-1} \cmidrule(lr){3-3}
      Total & & 79\,961 \\
      \bottomrule
    \end{tabular}
    \captionof{table}{Ground Truth statistics. \label{tab:gt2}}
  \end{minipage}
\end{minipage}
\end{figure}

The study site covers an area around the town of \textit{Koumbia}, in the Province of Tuy, \textit{Hauts-Bassins} region, in the south-west of Burkina Faso. This area has a surface of about 2338 $ km^2 $, and is situated in the sub-humid sudanian zone. The surface is covered mainly by natural savannah (herbaceous and shrubby) and forests, interleaved with a large portion of land (around 35\%) used for rainfed agricultural production (mostly smallholder farming). The main crops are cereals (maize, sorghum and millet) and cotton, followed by oleaginous and leguminous crops. Several temporary watercourses constitute the hydrographic network around the city of Koumbia.
Figure~\ref{fig:loc-koumbia} presents the study site with the reference data (ground truth) superposed on a Sentinel-2 image.

The satellite data consists of a time series of Sentinel-2 images spanning the year 2020 from January to December~\citep{Jolivot2021}. All images were provided by the THEIA Pole platform\footnote{\url{http://theia.cnes.fr}} at level-2A, which consist of atmospherically corrected surface reflectances (cf. MAJA processing chain~\citep{rs70302668}) and relative cloud/shadow masks. A standard pre-processing was performed over each band to replace cloudy pixel values as detected by the available cloud masks based on the method proposed in~\citep{jordi_inglada_2016_58150}. 
Finally, from the spectral raw bands at 10-m of spatial resolution the NDVI (Normalized Differential Vegetation Index) was derived.

The GT (ground truth) data for the study site is a collection of (i) digitized plots from a GPS field mission performed in October 2020 and mostly covering classes within cropland and (ii) additional reference plots on non-crop classes obtained by photo-interpretation by an expert. Finally, the polygons have been rasterized at the S2 spatial resolution (10-m), resulting in 79961 labeled pixels. The statistics related to the GT are reported in~\Cref{tab:gt2}.

\section{Proposed method} \label{sec:method}



\begin{minipage}[h]{0.44\textwidth}
\paragraph{Architecture overview}
The proposed GAN (generative adversarial network)-inspired architecture is shown in Fig.~\ref{fig:architecture}.
A counterfactual $x_{\mathrm{CF}}$ is obtained for each input sample $x$ by adding a perturbation $\delta$:
\begin{align}
    x_{\mathrm{CF}} = x + \delta
\end{align}
The perturbation $\delta$ is generated by a \emph{Noiser} module learned with the goal to swap the prediction of the \emph{Classifier}. A \emph{Discriminator} module is leveraged to ensure the generation of realistic counterfactual examples.  
\end{minipage}
\hfill
\begin{minipage}[h]{0.55\textwidth}
    \centering
    \includegraphics[width=0.99\textwidth]{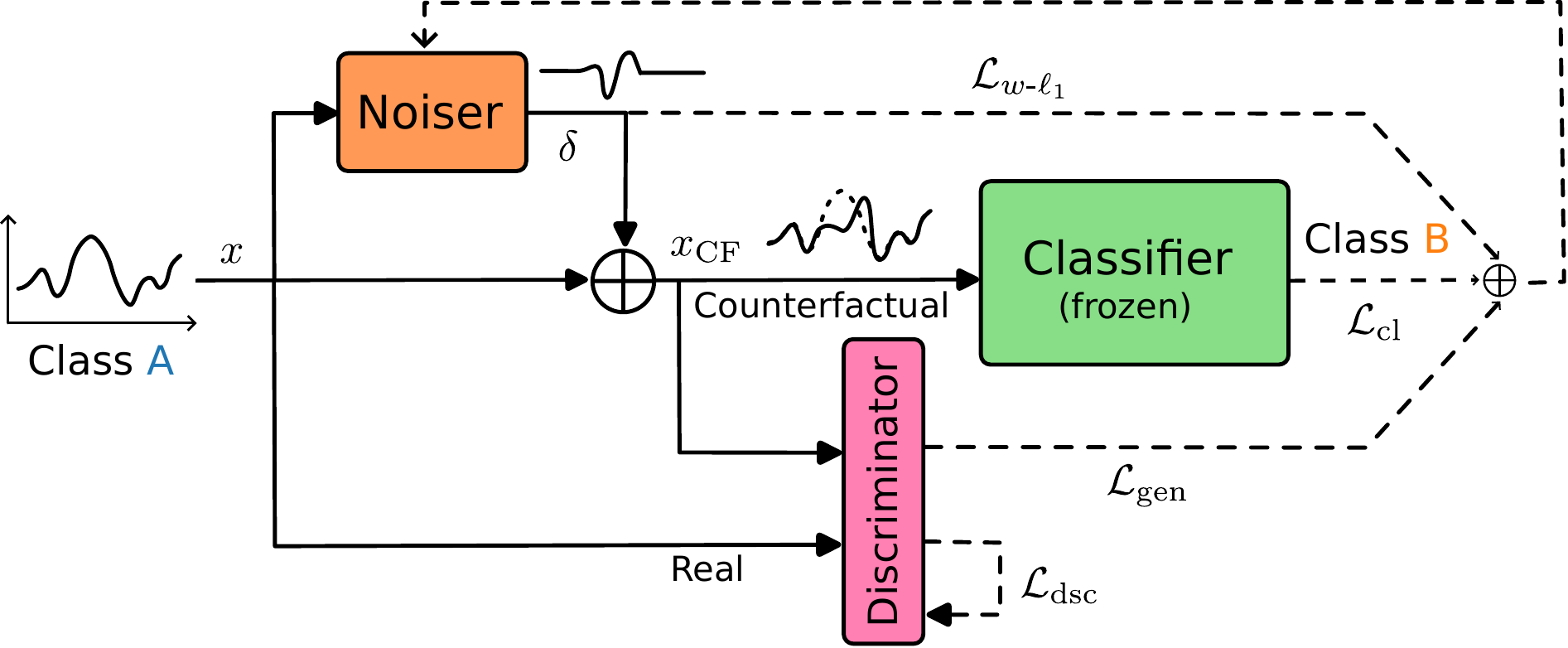}
    \captionof{figure}{Schematic view of the proposed approach. \label{fig:architecture}}
\end{minipage}

\paragraph{Implementation and training}

Regarding the different components of the proposed architecture, we get inspiration from state-of-the-art  satellite image time series land cover mapping literature. 
For the \textit{Classifier} network we leverage the Temporal Convolutional Neural Network (TempCNN) model \citep{PelletierWP19}. This architecture has an encoder based on several one-dimensional convolutional layers to explicitly cope with the temporal dimension of the time series data followed by two fully connected layers and a final output layer to provide the multi-class decision.
The same architecture is used for the \textit{Discriminator} module, replacing the output layer with a single neuron with sigmoid activation.
%
The \textit{Noiser} module is implemented as a multi-layer perceptron network with two hidden layers (each with 128 neurons) and an output layer with the same dimensionality of the time series data. For each of the hidden layers, batch normalization, tangent activation function and a drop-out regularization are employed in this order while for the output layer only the tangent activation function is used.
(to restrict the output domain between $\pm$1, like the NDVI index).
%
The \textit{Classifier} model is pre-trained on the training set and, then, frozen during the adversarial learning stage ---devoted to learn the \textit{Noiser} and \textit{Discriminator} modules (see \cref{eq:loss_gen}). 

The \textit{Noiser} module is updated w.r.t. a composite loss made of three parts detailed further below. 
\begin{align}
   \mathcal{L}_{\mathrm{noiser}} = \mathcal{L}_{\mathrm{cl}} + \lambda_{\mathrm{gen}}  \mathcal{L}_{\mathrm{gen}} +  \lambda_{w\text{-}\ell_1}  \mathcal{L}_{w\text{-}\ell_1}
\end{align}

\paragraph{Class-swapping loss} \label{ssec:loss_cl}

To generate counterfactuals that effectively change the predicted class we use the following loss
which enforces the reduction of the classifier's softmax output $p(y^{(i)})$ for the original label $y^{(i)}$, eventually leading to a change on the predicted class:
\begin{align}
   \mathcal{L}_{\mathrm{cl}} = - \frac{1}{n} \sum_{i=1}^n \log (1 - p(y^{(i)}))
\end{align}
Note that, conversely to standard literature~\citep{FilaliBoubrahimi2022,Lang2022} in which a target class for the counterfactual example is chosen a priori, here we purposely do not enforce the prediction of a predefined target class. 
Instead, we let the \emph{Noiser} free to generate a perturbation $\delta$ that will change the classifier output to any other class different from $y_i$.

\paragraph{GAN-based regularization for plausibility}  \label{ssec:loss_gan}

Counterfactual plausibility is enforced via a GAN-inspired architecture, where the \textit{Discriminator} is trained to identify unrealistic counterfactuals while, simultaneously, 
the \emph{Noiser} module acts as a generator with the goal to fool the discriminator in a two player game.
%
The following non-saturating GAN losses are used for the adversarial training: 
\begin{align} \label{eq:loss_gen}
   \mathcal{L}_{\mathrm{dsc}} = - \frac{1}{n} \sum_{i=1}^n \left[ \log   D(x^{(i)})  + \log  \left(1 - D(x_{\mathrm{CF}}^{(i)})\right) \right],
   & &
  \mathcal{L}_{\mathrm{gen}} = -\frac{1}{n} \sum_{i=1}^n  \log  \left(D(x_{\mathrm{CF}}^{(i)})\right)
\end{align}
where $D(x^{(i)})$ denotes the discriminator's output for a real input $x^{(i)}$ (with expected output 1) and $D(x_{\mathrm{CF}}^{(i)})$ its output for a fake input $x_{\mathrm{CF}}^{(i)}$ (with expected output 0). The loss $\mathcal{L}_{\mathrm{dsc}}$ is minimized by the \emph{Discriminator} while $\mathcal{L}_{\mathrm{gen}}$ is minimized the generator (i.e., the \emph{Noiser}).

\paragraph{Unimodal regularization for time-contiguity} \label{ssec:loss_uni} 

To generate perturbations concentrated around a contiguous time frame we employ a weighted L1-norm penalization, with weights growing quadratically around a central time $\tilde{t}^{(i)}$ chosen independently for each sample $i \in \{1, \dots, n\}$:
\begin{align} \label{eq:loss_uni}
    \mathcal{L}_{w\text{-}\ell_1} = \frac{1}{n} \sum^{n}_{i=1}  \sum^{T}_{t=1} d(t, \tilde{t}^{(i)})^2  |\delta^{(i)}_t|
\end{align}
where, for the $i$-th sample, $\tilde{t}^{(i)}$ is chosen as the time step with the highest absolute value perturbation $\tilde{t}^{(i)} = \argmax_t |\delta_t^{(i)}|$.
%
To avoid biasing $\tilde{t}$ towards the center, we use the modulo distance $d(t,\tilde{t}) = \min \left( (t-\tilde{t})\%T,  (\tilde{t}- t)\%T \right)$ which treats the time samples as a circular list.
Besides enforcing sparsity, penalizing the entries of $\delta$ also enforces the proximity (similarity) between $x_{\mathrm{CF}}$ and $x$.

\section{Results} \label{sec:results}

In this section, we analyse the class transitions induced by the counterfactual generation process and discuss some examples of generated counterfactual explanations. 
In the appendix, we also discuss per-class average perturbations (sec. \ref{ssec:avg_perturbation}), assess the plausibility of the generated counterfactual examples via an anomaly detection strategy (sec. \ref{ssec:inlier}) and perform an ablation analysis (sec. \ref{ssec:ablation}).


\paragraph{Experimental setup}

The \emph{Koumbia} study case described in Section~\ref{sec:data} was split into training, validation and test sets containing respectively 50-17-33\% of the 79961 samples. Each data sample corresponds to a (univariate) NDVI time series with 24 time samples. 
First, the \emph{Classifier} was trained over 1000 epochs with batch size 32 and Adam optimizer with learning rate $10^{-4}$ and weight decay of same value. The model weights corresponding to the best obtained F1-score on the validation set were kept.
Then, with the classifier weights frozen, the \emph{Noiser} and \emph{Discriminator} modules are simultaneously trained over 100 epochs with batch size 128 and Adam optimizer. 
Finally, we empirically set $\lambda_{\mathrm{gen}}=5\!\cdot\!10^{-1}$ and $\lambda_{w\text{-}\ell_1}=5\!\cdot\!10^{-2}$ on the reported results.

\vspace{-0.3cm}
\paragraph{Visualizing class relationships}

\begin{figure}[t]
\vspace{-0.7cm}
\centering
\includegraphics[width=.4\columnwidth]{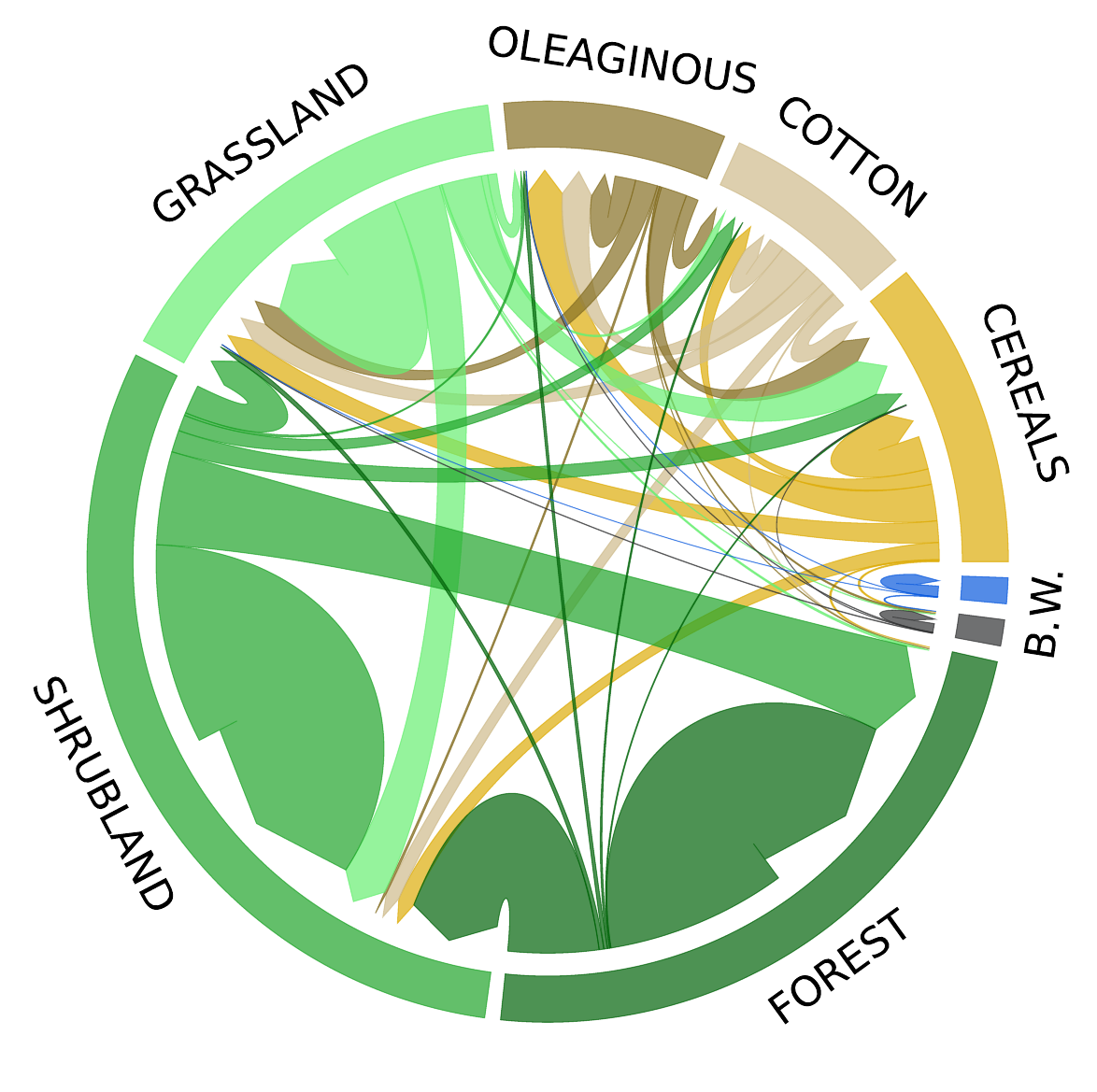}
\includegraphics[width=.4\columnwidth]{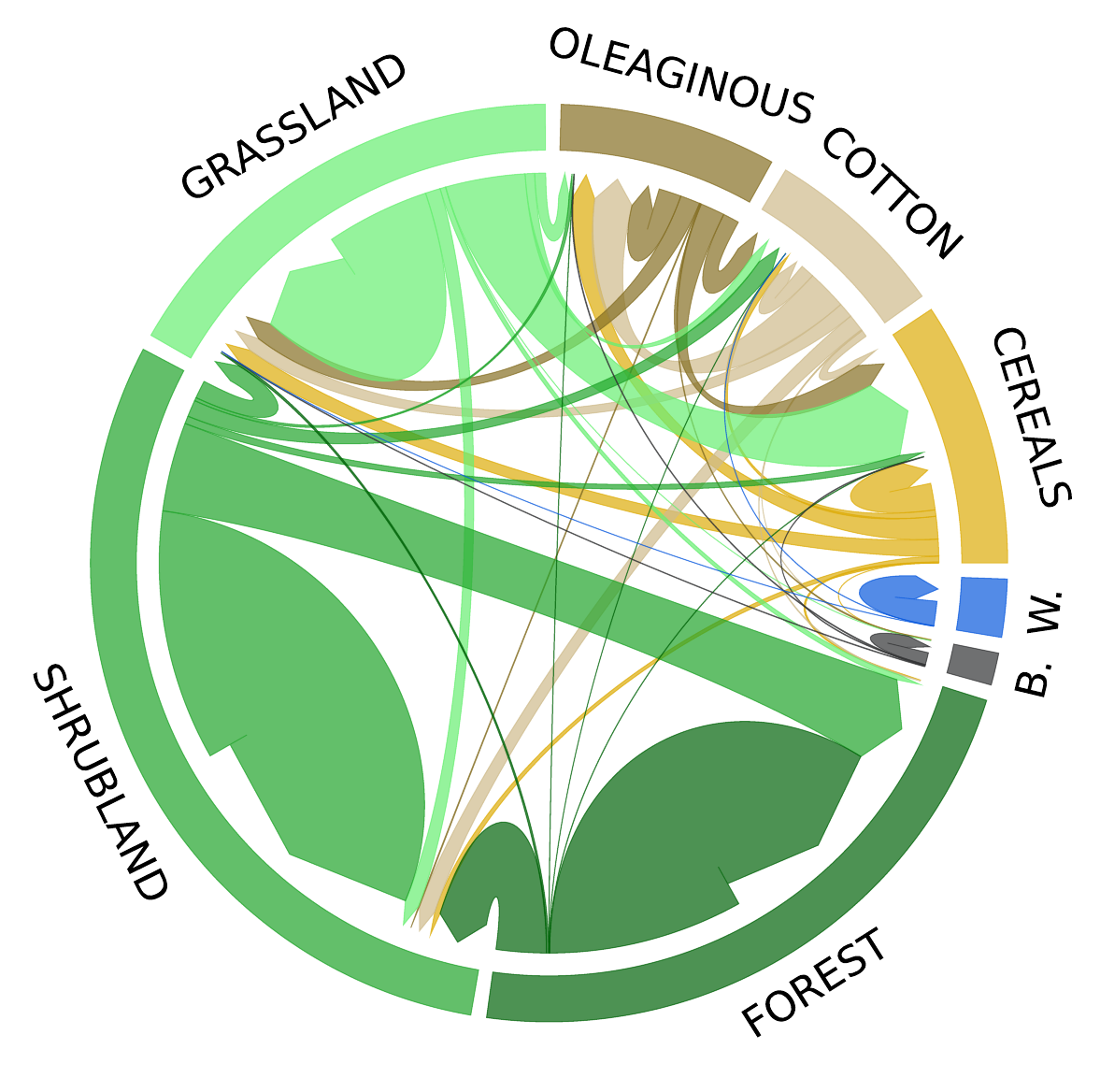}
\vspace{-0.5cm}
\caption{Summary of class transitions induced by the counterfactuals. Training data (left) and test data (right), where B. stands for \emph{Bare Soil} and W. for \emph{Water} classes.}
\label{fig:chord}
\end{figure}

Class transitions induced by the counterfactual samples are summarized in Fig.~\ref{fig:chord},
with arrow widths proportional to the number of counterfactuals leading to that specific transition.
The left (resp. right) graph was generated by feeding the network with each of the training (resp. test) data samples. They present very similar behavior, indicating that the proposed method generalizes well to unseen data. We recall that class transitions are not pre-defined; on the contrary, our method allows input samples to freely split-up into multiple target classes. 
Transitions obtained in such a way thus bring up valuable insights on the relation between classes. 

\vspace{-0.3cm}
\paragraph{Counterfactual examples}

\begin{figure}[t]
\vspace{-0.4cm}
\centering
\includegraphics[width=.38\columnwidth, trim={0 0 0.55cm 0}]{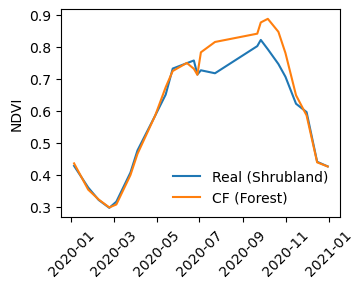} 
\includegraphics[width=.38\columnwidth,trim={0.55cm 0 0 0}, clip]{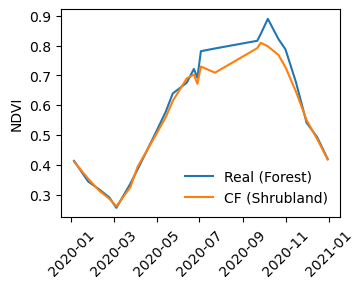} 
\vspace{-0.3cm}
\caption{Examples of original time series with corresponding counterfactual from classes \emph{Shrubland} (4) and \emph{Forest} (5) on both ways. \label{fig:examples}}
\vspace{-0.5cm}
\end{figure}

Two illustrative examples of counterfactual explanations are shown in  Fig.~\ref{fig:examples}.
It is interesting to observe the similarity between the generated counterfactual and a real data example from the same class (on the neighboring plot).
To transform a \emph{Shrubland} sample into a \emph{Forest} one, NDVI is added between the months of July and October. The opposite is done to obtain the reverse transition, which is reassuring.
%
One can verify that the obtained counterfactuals do look realistic besides differing from the real signal only on a contiguous time window. These two properties have been explicitly enforced via the losses in eqs. \eqref{eq:loss_gen} and \eqref{eq:loss_uni}.

\section{Conclusion} \label{sec:conclusion}
\vspace{-0.3cm}

In this paper, we presented a new framework to generate counterfactual SITS samples of vegetation indices (NDVI) for the land cover classification task. The proposed method overcomes the restriction to apriori define the source and the target classes for the counterfactual generation process while leveraging adversarial learning to ensure realistic counterfactual samples.
A possible future work would be to extend the framework to the case of multivariate time series satellite data.

\bibliography{refs}
\bibliographystyle{iclr2023_conference}

\newpage
\appendix
\section{Appendix}
\subsection{Average perturbation examples} \label{ssec:avg_perturbation}

\begin{figure}[h]
\centering
\includegraphics[width=.49\columnwidth, trim={0 0 0.32cm 0}]{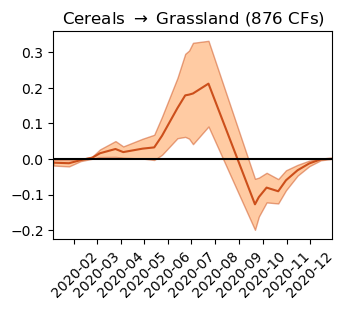} 
\includegraphics[width=.49\columnwidth, trim={0.32cm 0 0 0}, clip]{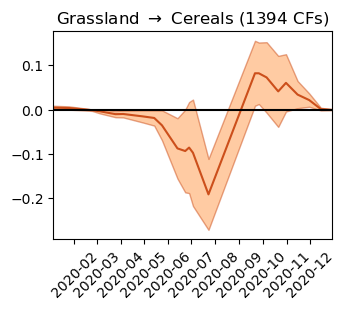} 
\caption{Examples of average counterfactual perturbations between classes \emph{Cereals} and \emph{Grassland} on both ways. Shaded area corresponds to the standard deviation. \label{fig:average}}
\end{figure}

Examples of average perturbation profiles for two different class transitions are depicted in Fig~ \ref{fig:average}. 
It is interesting to notice how the perturbations correspond roughly to the opposite of each other, which is quite suitable since they correspond to opposite transitions between the same two classes. 

\subsection{Plausibility analysis} \label{ssec:inlier}

\begin{figure}
    \includegraphics[width=.4\linewidth]{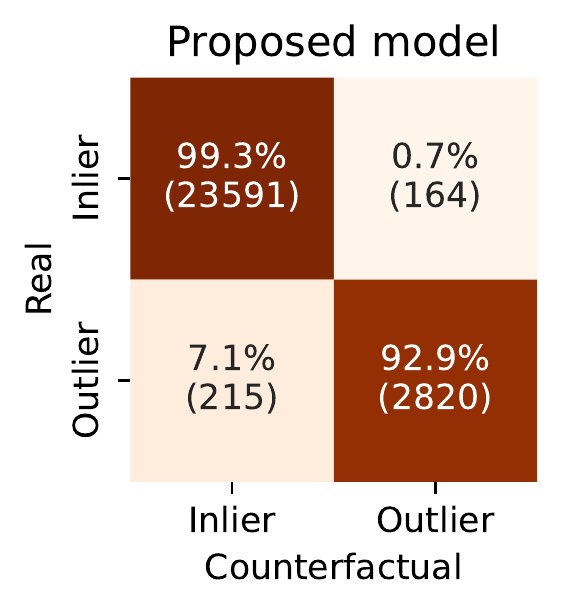}
    \includegraphics[width=.4\linewidth]{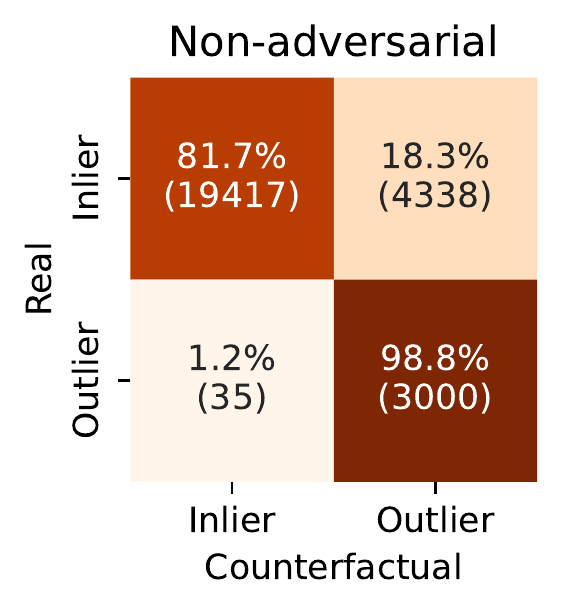}
    \caption{Isolation forest results on real (rows) and counterfactual data (columns). Proposed model with (left) and without (right) adversarial loss during training. Row-normalized percentages.}
    \label{fig:IF_eval}
\end{figure}

In this section, we quantify to what extent the proposed counterfactual explanations fit the original data distribution. To do so, we run an anomaly detection method, Isolation Forest \citep{Li2018}, on both the original data and corresponding counterfactuals. To attest the importance of the proposed adversarial training for the generation of realistic/plausible counterfactuals, we perform an ablation study confronting the proposed model trained with and without the generator loss in Eq.~\eqref{eq:loss_gen}. Fig.~\ref{fig:IF_eval} shows contingency matrices relating the isolation forest outputs on the original data (rows) and on the corresponding counterfactual explanations (columns). Two counterfactual generation approaches are investigated: the proposed method (left matrix) and its non-adversarial variant (right matrix). In the figures, diagonal entries correspond to matching isolation forest outputs --i.e., same prediction (inlier/outlier) for both real and counterfactual data. Later, in \Cref{tab:IF_evaluation} we compute some metrics on such contingency matrices to further quantify and summarize the behaviour of the compared methods. The proposed counterfactual model achieves impressive results, even leading to more samples identified as inliers than the real data itself (23806 against 23755), since proposed approach converts less inliers into outliers (164) than the other way around (215).

The non-adversarial variant, on the other hand, obtains considerably more degraded results, as it converts as many as 4338 real inlier samples into outliers (about 20 times more). 
Such a gap becomes evident when looking at the corresponding accuracy and normalized mutual information (NMI) computed w.r.t. the isolation forest results on the original data (cf. \Cref{tab:IF_evaluation}). Such scores measure to what degree the inlier/outlier partitioning obtained on the counterfactual samples (for each of the two compared variants) matches the one obtained on the original data. The higher they are the better the two partitions match. The  obtained results clearly show that counterfactual plausibility is achieved thanks to the adversarial training process.


\begin{table}[!htb]
\centering
 \begin{tabular}{cccc}
  \toprule
  Method &  Accuracy & NMI & Inliers ratio \\
  \cmidrule(lr){1-1} \cmidrule(lr){2-2} \cmidrule(lr){3-3} \cmidrule(lr){4-4}
  Proposed        &  \textbf{98.6\%} &  \textbf{0.808} & \textbf{88.9\%} \\
  Non-adversarial & 83.7\% & 0.337 & 72.6\%\\
  \bottomrule
 \end{tabular}
\caption{Plausibility analysis using different performance metrics. Isolation Forest results on the real data were used as ground truth for the accuracy and NMI scores. \label{tab:IF_evaluation}}
\end{table}

\subsection{Other ablation studies} \label{ssec:ablation}

In \Cref{tab:ablation} we compare the number of successful class-swapping counterfactual samples as well as the average $\ell_2$ and  $\ell_1$ norms of the perturbations $\delta$ generated by the proposed model and two variants ignoring the generator loss ($\mathcal{L}_{\mathrm{gen}}$) and the weighted-$\ell_1$ loss ($\mathcal{L}_{w\text{-}\ell_1}$), respectively.

One can see that the removal of the auxiliary losses significantly bumps the class-swapping rate, but it happens at the expense of either: 1) counterfactual plausibility, as shown in the Section~\ref{ssec:inlier} for the removal of  $\mathcal{L}_{\mathrm{gen}}$; 2) counterfactual proximity/similarity, as demonstrated by the dramatic increase on the norm of the generated perturbations (or, equivalently, the distance between $x$ and $x_{\mathrm{CF}}$) upon removal of $\mathcal{L}_{w\text{-}\ell_1}$.

\begin{table}[!htb]
\centering
 \begin{tabular}{cccc}
  \toprule
  Method &  Class-swap CF & Average $\|\delta\|_2$ & Average $\|\delta\|_1$ \\
  \cmidrule(lr){1-1} \cmidrule(lr){2-2} \cmidrule(lr){3-3} \cmidrule(lr){4-4}
  Proposed        &  43.8\% &  0.24 $\pm$ 0.18 & 0.76 $\pm$ 0.54 \\
  Without $\mathcal{L}_{\mathrm{gen}}$ & 83.7\% & 0.97 $\pm$ 0.47 & 1.69 $\pm$ 0.99 \\
  Without $\mathcal{L}_{w\text{-}\ell_1}$ & 99.6\% & 4.79 $\pm$ 0.07 & 23.3 $\pm$ 0.53 \\
  \bottomrule
 \end{tabular}
\caption{Ablation study on test data. \label{tab:ablation}}
\end{table}

\end{document}